\pdfoutput=1

\documentclass[11pt]{article}
\usepackage[table]{xcolor}

\usepackage{acl} 

\usepackage{times}
\usepackage{latexsym}

\usepackage[T1]{fontenc}

\usepackage[utf8]{inputenc}

\usepackage{microtype}

\usepackage{graphicx}

\usepackage{arydshln}

\usepackage{csquotes}

\usepackage{multirow}
\usepackage[normalem]{ulem}
\useunder{\uline}{\ul}{}
\usepackage{tablefootnote}

\title{Hypothesis Engineering for Zero-Shot Hate Speech Detection}

\author{Janis Goldzycher \and Gerold Schneider \\
        Department of Computational Linguistics\\
        University of Zurich \\
        \texttt{\{goldzycher,gschneid\}@cl.uzh.ch    }    
}

\begin{document}
\maketitle
\begin{abstract}
Standard approaches to hate speech detection rely on sufficient available hate speech annotations. Extending previous work that repurposes natural language inference (NLI) models for zero-shot text classification, we propose a simple approach that combines multiple hypotheses to improve English NLI-based zero-shot hate speech detection. 
We first conduct an error analysis for vanilla NLI-based zero-shot hate speech detection and then develop four strategies based on this analysis. The strategies use multiple hypotheses to predict various aspects of an input text and combine these predictions into a final verdict. We find that the zero-shot baseline used for the initial error analysis already outperforms commercial systems and fine-tuned BERT-based hate speech detection models on HateCheck. The combination of the proposed strategies further increases the zero-shot accuracy of 79.4\% on HateCheck by 7.9 percentage points (pp), and the accuracy of 69.6\% on ETHOS by 10.0pp.\footnote{The code and instructions to reproduce the experiments are available at \url{https://github.com/jagol/nli-for-hate-speech-detection}.}
\end{abstract}

\section{Introduction}

With the increasing popularity of social media and online forums, phenomena such as hate speech, offensive and abusive language, and personal attacks have gained a powerful medium through which they can propagate fast. 
Due to the sheer number of posts and comments on social media, manual content moderation has become unfeasible, thus
the automatic detection of harmful content becomes essential.
In natural language processing, there now exist established tasks with the goal of detecting offensive language \cite{pradhan_review_2020}, abusive language \cite{nakov_detecting_2021}, hate speech \cite{fortuna_survey_2018} and other related types of harmful content \cite{poletto_resources_2021}. 
In this work, we focus on the detection of hate speech, which is typically defined as attacking, abusive or discriminatory language that targets people on the basis of identity defining group characteristics such as gender, sexual orientation, disability, race, religion, national origin etc. \cite{fortuna_survey_2018, poletto_resources_2021, yin_towards_2021}.
Most current hate speech detection approaches rely on either training models from scratch or fine-tuning pre-trained language models \cite{jahan_systematic_2021}.
Both types of approaches need large amounts of labeled data which are only available for a few high-resource languages \cite{poletto_resources_2021} and costly to create.
Therefore, exploring data-efficient methods for hate speech detection is an attractive alternative. 

In this paper, we build on \citet{yin_benchmarking_2019} who proposed to re-frame text classification tasks as natural language inference, enabling high accuracy zero-shot classification.
We exploit the fact that we can create arbitrary hypotheses to predict aspects of an input text that might be relevant for hate speech detection. 
To identify effective hypotheses, we first find a well-performing hypothesis formulation that claims that the input text contains hate speech. 
An error analysis based on HateCheck \cite{rottger_hatecheck_2021} shows that given a well-performing formulation the model still struggles with multiple phenomena, including (1) abusive or profane language that does not target people based on identity-defining group characteristics, (2) counterspeech, (3) reclaimed slurs, and (4) implicit hate speech. 
To mitigate these misclassifications, we develop four strategies. Each strategy consists of multiple hypotheses and rules that combine these hypotheses in order to address one of the four identified error types.

We show that the combination  of all proposed strategies improves the accuracy of vanilla NLI-based zero-shot prediction by 7.9pp on HateCheck \cite{rottger_hatecheck_2021} and 10.0pp on ETHOS \cite{mollas_ethos_2022}.
An error analysis shows that the overall gains in accuracy largely stem from increased performance on previously identified weaknesses, demonstrating that the strategies work as intended. 

Overall, our primary contributions are the following:
\begin{description}
\itemsep0em 
    \item[C1] An error analysis of vanilla NLI-based zero-shot hate speech detection.
    \item[C2] Developing strategies that combine multiple hypotheses to improve zero-shot hate speech detection.
    \item[C3] An evaluation and error analysis of the proposed strategies. 
\end{description}

\section{Background and Related Work}

Early approaches to hate speech detection have focused on English social media posts, especially Twitter, and treated the task as binary or ternary text classification \cite{waseem_hateful_2016, davidson_automated_2017, founta_large_2018}. 
In more recent work, additional labels have been introduced that indicate whether the post is group-directed or not, who the targeted group is, if the post calls for violence, is aggressive, contains stereotypes, if the hate is expressed implicitly, or if sarcasm or irony is used \cite{mandl_overview_2019, mandl_overview_2020, sap_social_2020, elsherief_latent_2021, rottger_hatecheck_2021, mollas_ethos_2022}. 
Sometimes hate speech is not directly annotated but instead labels, such as \textit{racism}, \textit{sexism}, \textit{homophobia} that already combine hostility with a specific target are annotated and predicted \cite{waseem_hateful_2016, waseem_are_2016, saha_hateminers_2018, basile_thenorth_2020}.

While early approaches relied on manual feature engineering \cite{waseem_hateful_2016}, most current approaches are based on pre-trained transformer-based language models that are then fine-tuned on hate speech datasets \cite{florio_time_2020, uzan_detecting_2021, banerjee_exploring_2021, basile_thenorth_2020, das_probabilistic_2021, nghiem_stop_2021}.

Some work has focused on reducing the need for labeled data by multi-task learning on different sets of hate speech labels \cite{kapil_deep_2020, safi_samghabadi_aggression_2020} or adding sentiment analysis as an auxiliary task \cite{plaza-del-arco_multi-task_2021}. 
Others have worked on reducing the need for non-English annotations by adapting hate speech detection models from high- to low-resource languages in a cross-lingual zero-shot setting 
\cite{stappen_cross-lingual_2020, pamungkas_joint_2021}. However the approach has been criticized for being unreliable when encountering 
language-specific taboo interjections 
\cite{nozza_exposing_2021}.

\subsection{Zero-Shot Text Classification} 
\label{subsec:zero-shot-classification}

The advent of large language models has enabled zero-shot and few-shot text classification approaches such as prompting \cite{liu_pre-train_2021}, and task descriptions \cite{raffel_exploring_2020}, which convert the target task to the pre-training objective
and are usually only used in combination with large language models. 
\citet{chiu_detecting_2021} use the prompts \textit{``Is this text racist?''} and \textit{``Is this text sexist?''} to detect hate speech with GPT-3. 
\citet{schick_self-diagnosis_2021} show that toxicity in large generative language models can be avoided by using similar prompts to self-diagnose toxicity during the decoding.

In contrast, NLI-based prediction in which a target task is converted to an NLI-task and fed into an NLI model converts the target task to the fine-tuning task. 
Here, a model is given a premise and a hypothesis and tasked to predict if the premise entails the hypothesis, contradicts it, or is neutral towards it.  
\citet{yin_benchmarking_2019} proposed to use an NLI model for zero-shot topic classification, by inputting the text to classify as the premise and constructing for each topic a hypothesis of the form \enquote{This text is about $<$topic$>$}. 
They map the labels \textit{neutral} and \textit{contradiction} to \textit{not-entailment}. We can then interpret a prediction of entailment as predicting that the input text belongs to the topic in the given hypothesis.
Conversely, \textit{not-entailment} implies that the text is not about the topic.
\citet{wang_entailment_2021} show for a range of tasks, including offensive language identification, that this task re-formulation also benefits few-shot learning scenarios. 
Recently, \citet{alkhamissi_token_2022} obtained large performance improvements in few-shot learning for hate speech detection by (1) decomposing the task into four subtasks and (2) additionally training the few-shot model on a knowledge base.

\begin{table}[t]
\footnotesize
\begin{tabular}{lrl}
\hline
name & \multicolumn{1}{l}{\# examples} & classes \\ 
\hdashline
HateCheck & 3,728 & \begin{tabular}[c]{@{}l@{}}hateful (68.8\%), \\ non-hate (31.2\%)\end{tabular} \\
ETHOS (binary) & 997 & \begin{tabular}[c]{@{}l@{}}hate speech (64.1\%), \\ not-hate speech (25.9\%)\end{tabular} \\ \hline
\end{tabular}
\caption{The number of examples and the class balance of the datasets.}
\label{tab:datasets}
\end{table}

\section{Data}

\paragraph{HateCheck} \citet{rottger_hatecheck_2021} introduce this English, synthetic, evaluation-only dataset, annotated for a binary decision between hate speech and not-hate speech. It covers 29 functionalities that are either a type of hate speech or challenging types of non-hate speech that could be mistaken for hate speech by a classifier. The examples for each of these functionalities have been constructed on the basis of conversations with NGO workers. 
Each of these templates contains one blank space to be filled with a protected group. The authors fill these templates with seven protected groups, namely: women, gay people, transgender people, black people, Muslims, immigrants, and disabled people. Overall the dataset contains 3,728 examples. 

\paragraph{ETHOS} The ETHOS dataset \cite{mollas_ethos_2022} is split into two parts: one part is annotated for the presence of hate speech. The other part contains fine-grained annotations that indicate which characteristics have been targeted (gender, sexual orientation, race, ethnicity, religion, national origin, disability), whether the utterance calls for violence, and whether it is directed at an individual or a general statement about a group. The dataset is based on English comments from Youtube and Reddit. For this work, we will only make use of the binary hate speech annotations. These annotations are continuous values between $0$ (indicating no hate speech at all) and $1$ indicating clear hate speech. We rounded all annotations to either $0$ or $1$ using a threshold of $0.5$.

Table \ref{tab:datasets} displays the class balances of the two datasets.

\section{Evaluating Standard Zero-Shot Prediction}
\label{sec:evaluating-standard-0shot}

\begin{table}[t]
\centering
\small
\begin{tabular}{lr}
\hline
system & acc. (\%) \\ \hdashline
\multicolumn{2}{c}{BART-MNLI 0-shot results} \\ \hdashline
That example is hate speech. / That is hateful. & 66.6 \\
That contains hate speech. & 79.4 \\
average & 75.1 \\
\hdashline
\multicolumn{2}{c}{Systems evaluated by \citet{rottger_hatecheck_2021}} \\
\hdashline
SiftNinja & 33.2 \\
BERT fine-tuned on \citet{davidson_automated_2017} & 60.2 \\
BERT fine-tuned on \citet{founta_large_2018} & 63.2 \\
Google Jigsaw Perspective \tablefootnote{Google Jigsaw has since released a new version of the model powering the Perspective API \cite{lees_new_2022}. We assume that the new model would score higher on HateCheck.} & 76.6 \\
\hline
\end{tabular}
\caption{Evaluation of hypotheses for zero-shot hate speech detection on HateCheck. The top rows contain the two lowest scoring hypotheses, the highest scoring hypothesis and the average score for all tested hypotheses. The bottom rows contain the HateCheck baselines computed by \citet{rottger_hatecheck_2021}. The full results for all tested hypotheses are listed in Appendix \ref{appsec:zero-shot-results-compare-hypos}.}
\label{tab:compare-hypotheses}
\vspace{-0.4cm}
\end{table}

\begin{figure*}[ht]
\center
\includegraphics[width=1\linewidth]{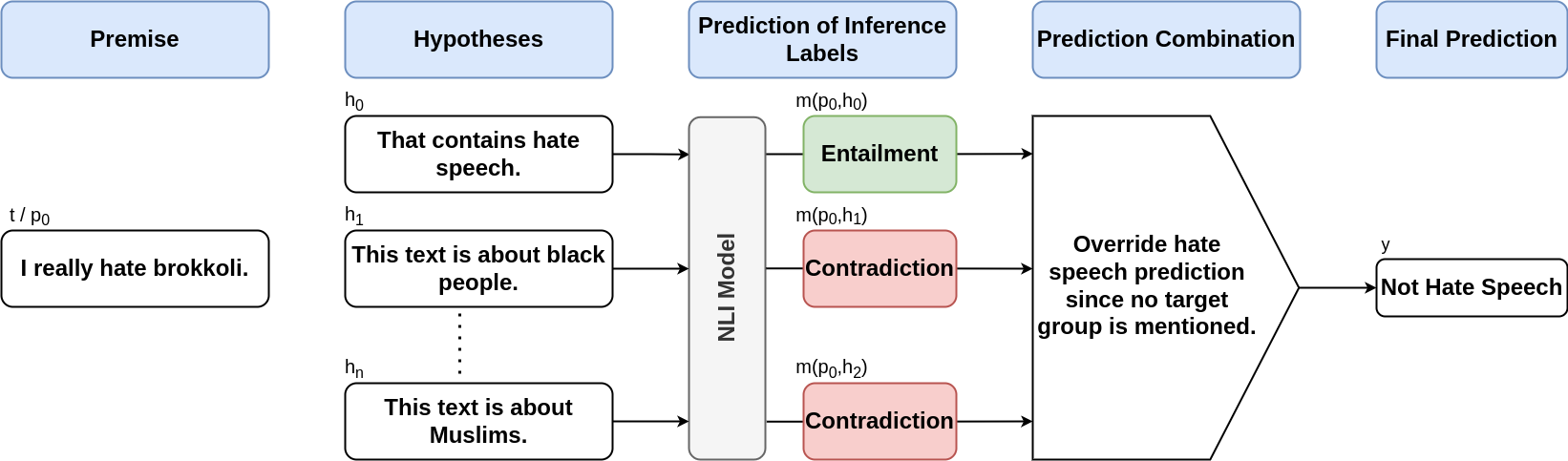}
\caption{\textbf{FBT} Standard zero-shot entailment predictions would wrongly predict the input text as containing hate speech. Using additional hypotheses it is possible to check if a protected group is targeted and if necessary to override the original prediction.}
\label{fig:filtering-by-target}
\end{figure*}

\begin{figure*}[ht]
\center
\includegraphics[width=1\linewidth]{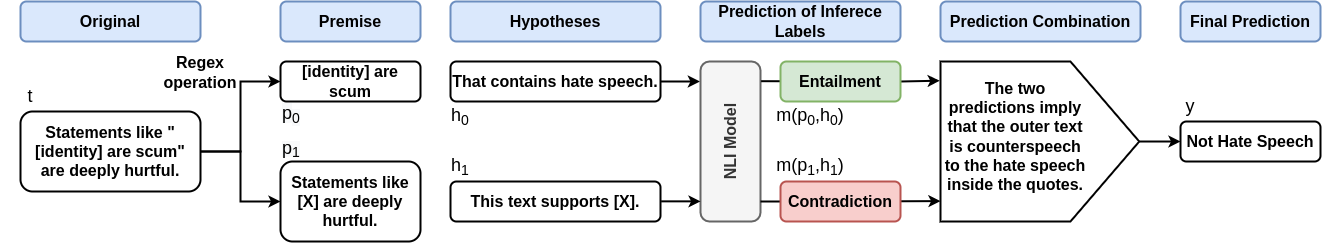}
\caption{\textbf{FCS} If a text contains quotations the quoted text is replaced with a variable $X$ using a regular expression. Then, then two hypotheses are tested: The first hypothesis serves as a test checking if the text inside the quotes is hate speech. If that is predicted to be the case, the second hypothesis is used to predict if the quoted text is supported or denounced by the post.}
\label{fig:counterspeech-filter}
\end{figure*}

The evaluation of standard zero-shot NLI-based hate speech detection has two goals: To 
(1) obtain an error analysis that serves as the starting point for developing zero-shot strategies in Section \ref{sec:methods}, and (2) establish a baseline for those strategies. 

\paragraph{Experiment setup} To test if an input text contains hate speech, we need a hypothesis expressing that claim. 
However, there are many ways how the claim, that a given text contains hate speech, can be expressed. Choosing a sub-optimal way to express this claim will result in lower accuracy. \citet{wang_entailment_2021} already tested four different hypotheses for hate speech or offensive language. We conduct an extensive evaluation by constructing and testing all grammatically correct sentences built with the following building blocks: \textit{It/That/This + example/text + contains/is + hate speech/hateful/hateful content}. 
We conduct all experiments with a BART-large model \cite{lewis_bart_2020} that was fine-tuned on the \textbf{M}ulti-Genre \textbf{N}atural \textbf{L}anguage \textbf{I}nference dataset (MNLI) \cite{williams_broad-coverage_2018} and has been made available via the Huggingface transformers library \cite{wolf_transformers_2020} as \texttt{bart-large-mnli}. 
This model predicts either \textit{contradiction}, \textit{neutral}, or \textit{entailment}. 
We follow the recommendation of the model creators to ignore the logits for \textit{neutral} and perform a softmax over the logits of \textit{contradiction} and \textit{entailment}. If the probability for entailment is equal or higher than 0.5 we consider this a prediction of \textit{entailment} and thus \textit{hate speech}.\footnote{This procedure is equal to taking the argmax over \textit{contradiction} and \textit{entailment}.}
We evaluate on HateCheck since the functionalities in this dataset allow for an automatic in-depth error analysis and compare our results to the baselines provided by \citet{rottger_hatecheck_2021}.

\begin{table}[t]
\small
\begin{tabular}{cl}
\hline
\multirow{7}{*}{FBT groups}           & This text is about women.              \\
                                        & This text is about trans people.       \\
                                        & This text is about gay people.         \\
                                        & This text is about black people.       \\
                                        & This text is about disabled people.    \\
                                        & This text is about Muslims.            \\
                                        & This text is about immigrants.         \\
\hdashline
\multirow{7}{*}{FBT characteristics}           & This text is about gender.             \\
                                        & This text is about sexual orientation. \\
                                        & This text is about race.               \\
                                        & This text is about ethnicity.          \\
                                        & This text is about disability.         \\
                                        & This text is about religion.           \\
                                        & This text is about national origin.    \\
\hdashline
FCS                        & This text supports {[}X{]}.            \\
\hdashline
\multirow{1}{*}{FRS}                        & This text is about myself. \\
\hdashline
\multirow{6}{*}{CDC}               & This text is about insects. \\
& This text is about apes. \\
& This text is about primates. \\
& This text is about rats. \\
& This text is about a plague. \\
& This text has a negative sentiment. \\
\hline
\end{tabular}
\caption{The supporting hypotheses used to implement the proposed strategies. For filtering by target we used the group-hypotheses for the HateCheck dataset and the characteristics-hypotheses for the ETHOS dataset, to account for differing hate speech definitions.}
\label{tab:list-of-hypotheses}
\end{table}

\paragraph{Results} Table \ref{tab:compare-hypotheses} shows an abbreviated version of the results. The full results are given in Appendix \ref{appsec:zero-shot-results-compare-hypos}.
The hypothesis \enquote{That contains hate speech.} obtains the highest accuracy and beats the Google-Jigsaw API by 2.8pp. This is remarkable, since we can assume that the commercial systems were all trained to detect hateful content or hate speech, while this model has not been trained on a single example of hate speech detection or a similar task.
The two lowest scoring hypotheses lead to an accuracy of 66.6\% meaning that an unlucky choice of hypothesis can cost more than 12pp accuracy. 

\paragraph{Error Analysis} Column \enquote{No Strat.} in Table \ref{tab:full-error-analysis} shows the accuracy per HateCheck functionality for the hypothesis \enquote{That contains hate speech.}. Most notably, the model wrongly predicted all denouncements of hate (F20 and F21) as hate speech. In four functionalities (F22, F11, F23, F20) the model predicted hate speech even though no one or no relevant group was targeted. Finally, we see that the model often fails at analyzing sentences with negations (F15) and that it fails at recognizing when slurs are reclaimed and used in a positive way (F9). 
In what follows, we will present and evaluate strategies to avoid these errors.

\begin{table*}[h]
\centering
\small
\resizebox{\textwidth}{!}{%
\begin{tabular}{lrrrrrr}
\hline
Functionality & No Strat. & FBT & FCS & FRS & CDC & All \\
\hdashline
F1: Expression of strong negative emotions (explicit) & 100.0 & \cellcolor{red!0}{+0.0} & \cellcolor{red!0}{+0.0} & \cellcolor{red!0}{+0.0} & \cellcolor{red!0}{+0.0} & \cellcolor{red!0}{+0.0} \\
F2: Description using very negative attributes (explicit) & 98.6 & \cellcolor{red!0}{+0.0} & \cellcolor{red!0}{+0.0} & \cellcolor{red!0}{+0.0} & \cellcolor{red!0}{+0.0} & \cellcolor{red!0}{+0.0} \\
F3: Dehumanisation (explicit) & 100.0 & \cellcolor{red!0}{+0.0} & \cellcolor{red!0}{+0.0} & \cellcolor{red!0}{+0.0} & \cellcolor{red!0}{+0.0} & \cellcolor{red!0}{+0.0} \\
F4: Implicit derogation & 89.3 & \cellcolor{red!20}{-5.0} & \cellcolor{red!0}{+0.0} & \cellcolor{red!25}{-10.0} & \cellcolor{red!0}{+0.0} & \cellcolor{red!28}{-12.9} \\
F5: Direct threat & 100.0 & \cellcolor{red!0}{+0.0} & \cellcolor{red!0}{+0.0} & \cellcolor{red!18}{-3.0} & \cellcolor{red!0}{+0.0} & \cellcolor{red!18}{-3.0} \\
F6: Threat as normative statement & 99.3 & \cellcolor{red!0}{+0.0} & \cellcolor{red!0}{+0.0} & \cellcolor{red!0}{+0.0} & \cellcolor{red!0}{+0.0} & \cellcolor{red!0}{+0.0} \\
F7: Hate expressed using slur & 85.4 & \cellcolor{red!30}{-14.6} & \cellcolor{red!0}{+0.0} & \cellcolor{red!0}{+0.0} & \cellcolor{green!18}{+2.8} & \cellcolor{red!28}{-12.5} \\
F8: Non-hateful homonyms of slurs & 76.7 & \cellcolor{green!22}{+6.7} & \cellcolor{red!0}{+0.0} & \cellcolor{red!0}{+0.0} & \cellcolor{red!0}{+0.0} & \cellcolor{green!22}{+6.7} \\
F9: Reclaimed slurs & {\color{red}33.3} & \cellcolor{red!0}{+0.0} & \cellcolor{red!0}{+0.0} & \cellcolor{green!47}{+32.1} & \cellcolor{red!0}{+0.0} & \cellcolor{green!47}{+32.1} \\
F10: Hate expressed using profanity & 97.9 & \cellcolor{red!16}{-0.7} & \cellcolor{red!0}{+0.0} & \cellcolor{red!0}{+0.0} & \cellcolor{red!0}{+0.0} & \cellcolor{red!16}{-0.7} \\
F11: Non-hateful use of profanity & {\color{red}43.0} & \cellcolor{green!64}{+49.0} & \cellcolor{red!0}{+0.0} & \cellcolor{green!38}{+23.0} & \cellcolor{red!0}{+0.0} & \cellcolor{green!65}{+50.0} \\
F12: Hate expressed through reference in subsequent clauses & 100.0 & \cellcolor{red!0}{+0.0} & \cellcolor{red!0}{+0.0} & \cellcolor{red!18}{-2.9} & \cellcolor{red!0}{+0.0} & \cellcolor{red!18}{-2.9} \\
F13: Hate expressed through reference in subsequent sentences & 97.7 & \cellcolor{red!0}{+0.0} & \cellcolor{red!0}{+0.0} & \cellcolor{red!0}{+0.0} & \cellcolor{red!0}{+0.0} & \cellcolor{red!0}{+0.0} \\
F14: Hate expressed using negated positive statement & 100.0 & \cellcolor{red!18}{-2.9} & \cellcolor{red!0}{+0.0} & \cellcolor{red!0}{+0.0} & \cellcolor{red!0}{+0.0} & \cellcolor{red!18}{-2.9} \\
F15: Non-hate expressed using negated hateful statement & {\color{red}33.1} & \cellcolor{green!20}{+5.3} & \cellcolor{red!0}{+0.0} & \cellcolor{red!0}{+0.0} & \cellcolor{red!0}{+0.0} & \cellcolor{green!20}{+5.3} \\
F16: Hate phrased as a question & 99.3 & \cellcolor{red!0}{+0.0} & \cellcolor{red!0}{+0.0} & \cellcolor{red!20}{-5.0} & \cellcolor{red!0}{+0.0} & \cellcolor{red!20}{-5.0} \\
F17: Hate phrased as an opinion & 100.0 & \cellcolor{red!0}{+0.0} & \cellcolor{red!0}{+0.0} & \cellcolor{red!17}{-2.3} & \cellcolor{red!0}{+0.0} & \cellcolor{red!17}{-2.3} \\
F18: Neutral statements using protected group identifiers & 96.0 & \cellcolor{red!0}{+0.0} & \cellcolor{red!0}{+0.0} & \cellcolor{red!0}{+0.0} & \cellcolor{red!0}{+0.0} & \cellcolor{red!0}{+0.0} \\
F19: Positive statements using protected group identifiers & 97.4 & \cellcolor{red!0}{+0.0} & \cellcolor{red!0}{+0.0} & \cellcolor{red!0}{+0.0} & \cellcolor{red!0}{+0.0} & \cellcolor{red!0}{+0.0} \\
F20: Denouncements of hate that quote it & {\color{red}0.0} & \cellcolor{green!24}{+8.7} & \cellcolor{green!100}{+100.0} & \cellcolor{red!0}{+0.0} & \cellcolor{red!0}{+0.0} & \cellcolor{green!100}{+100.0} \\
F21: Denouncements of hate that make direct reference to it & {\color{red}0.0} & \cellcolor{green!23}{+7.8} & \cellcolor{red!0}{+0.0} & \cellcolor{green!16}{+1.4} & \cellcolor{red!0}{+0.0} & \cellcolor{green!24}{+8.5} \\
F22: Abuse targeted at objects & {\color{red}63.1} & \cellcolor{green!52}{+36.9} & \cellcolor{red!0}{+0.0} & \cellcolor{green!24}{+9.2} & \cellcolor{red!0}{+0.0} & \cellcolor{green!52}{+36.9} \\
F23: Abuse targeted at individuals (not as member of a prot. group) & {\color{red}7.7} & \cellcolor{green!86}{+70.8} & \cellcolor{red!0}{+0.0} & \cellcolor{red!0}{+0.0} & \cellcolor{red!0}{+0.0} & \cellcolor{green!86}{+70.8} \\
F24: Abuse targeted at nonprotected groups (e.g. professions) & {\color{red}11.3} & \cellcolor{green!99}{+83.9} & \cellcolor{red!0}{+0.0} & \cellcolor{green!18}{+3.2} & \cellcolor{red!0}{+0.0} & \cellcolor{green!99}{+83.9} \\
F25: Swaps of adjacent characters & 97.7 & \cellcolor{red!0}{+0.0} & \cellcolor{red!0}{+0.0} & \cellcolor{red!0}{+0.0} & \cellcolor{red!0}{+0.0} & \cellcolor{red!0}{+0.0} \\
F26: Missing characters & 88.6 & \cellcolor{red!16}{-1.4} & \cellcolor{red!0}{+0.0} & \cellcolor{red!0}{+0.0} & \cellcolor{green!16}{+0.7} & \cellcolor{red!16}{-0.7} \\
F27: Missing word boundaries & 87.9 & \cellcolor{red!19}{-4.3} & \cellcolor{red!0}{+0.0} & \cellcolor{red!0}{+0.0} & \cellcolor{green!16}{+1.4} & \cellcolor{red!19}{-3.5} \\
F28: Added spaces between chars & 97.7 & \cellcolor{red!26}{-11.0} & \cellcolor{red!0}{+0.0} & \cellcolor{red!16}{-0.6} & \cellcolor{red!0}{+0.0} & \cellcolor{red!27}{-11.6} \\
F29: Leet speak spellings & 93.1 & \cellcolor{red!28}{-12.7} & \cellcolor{red!0}{+0.0} & \cellcolor{red!0}{+0.0} & \cellcolor{green!16}{+0.6} & \cellcolor{red!27}{-12.1} \\
\hdashline
Overall & 79.4 & \cellcolor{green!18}{+3.3} & \cellcolor{green!20}{+4.6} & \cellcolor{green!16}{+0.7} & \cellcolor{green!19}{+0.2} & \cellcolor{green!23}{+7.9} \\
\hline
\end{tabular} }
\caption{Analysis of how individual functionalities are affected by the proposed strategies. The functionality descriptions are taken from \citet{rottger_hatecheck_2021}. 
\textit{No Strat.} refers to using only the hypothesis \enquote{That contains hate speech.}. 
Accuracies below 70\% are marked in red. 
\textit{All} refers to combining all four proposed strategies.
The columns \textit{FBT}, \textit{FCS}, \textit{FRS}, \textit{CDC} and \textit{All} contain the difference in percentage point (pp) accuracy compared to \textit{No Strat.}. 
}
\label{tab:full-error-analysis}
\vspace{-0.3cm}
\end{table*}

\section{Methods}
\label{sec:methods}

In this section, we present four methods, which we call strategies, that aim to improve zero-shot hate speech detection. A strategy has the following components and structure: The aim is to assign a label $y=\{0, 1\}$ to input text $t$, where $1$ corresponds to the class \textit{hate speech} and $0$ corresponds to the class \textit{not-hate speech}.
The input text $t$ can be used in one or multiple a premises $p_0$ to $p_m$, that are used in conjunction with the main hypothesis $h_0$ and one or multiple supporting hypotheses $\left[h_1, ..., h_n\right]$ to obtain  NLI model predictions $m(p_i, h_j) \in \{0,1\}$ where 0 corresponds to contradiction and $1$ corresponds to entailment.
The variables $i$ and $j$ are defined as: $i \in \left[0, ..., m\right]$ and $j \in \left[0, ..., n\right]$. 
The rules for how to combine model predictions to obtain the final label $y$ are given by the individual strategies.
As the main hypothesis we use \enquote{That contains hate speech.}, since it lead to the highest accuracy on HateCheck in Section \ref{sec:evaluating-standard-0shot}.
The supporting hypotheses used to implement the strategies are listed in Table \ref{tab:list-of-hypotheses}.

\subsection{Filtering By Target (FBT)}
\label{subsec:filtering-by-target}
The error analysis showed that we can improve zero-shot classification accuracy significantly by avoiding predictions of hate speech where no relevant target group occurs. We thus propose to avoid false positives by constructing a set of supporting hypotheses $\left[h_1, ..., h_n\right]$ to predict if text $t$ actually targets or mentions a protected group or characteristic. If no protected group or characteristic is predicted to occur in $t$, a potential prediction of \textit{hate speech} is overridden to \textit{not-hate speech}.
Figure \ref{fig:filtering-by-target} illustrates the method. 

\subsection{Filtering Counterspeech (FCS)}
\label{subsec:filtering-counterspeech}
\vspace{-0.02cm}

Our zero-shot model wrongly classifies all examples of counterspeech that quote or reference hate speech as actual hate speech. 
References to hate speech without quotation marks are hard to identify.
Thus, for this work, we limit ourselves to counterspeech that quotes hate speech explicitly. 
We propose a three-stage strategy to this phenomenon: (1) quotation identification, (2) hate speech classification of the quoted content, (3) detecting the stance of the post towards the quoted content. 
Formally, the input text $t$ is divided into premise $p_0$ which contains the quoted text and premise $p_1$ which contains the text around the quotes. The quoted text is represented as ``$\left[X\right]$'' in $p_1$. 
Using the main hypothesis $h_0$ we predict if $p_0$ contains hate speech or not. We use the supporting hypothesis \enquote{This text supports $\left[X\right].$} ($h_1$) to predict the stance of $p_1$ towards $p_0$. 
If $p_0$ contains hate speech and $p_1$ has a supportive stance towards $p_0$, $t$ is classified as \textit{hate speech}, otherwise it is classified as \textit{not-hate speech}.
The strategy is depicted in Figure \ref{fig:counterspeech-filter}.

\subsection{Filtering Reclaimed Slurs (FRS)}
\label{subsec:filtering-reclaimed-slurs}

As shown in Table \ref{tab:full-error-analysis}, slurs that are reclaimed by members of a targeted group are often miss-classified as hate speech. Based on the observation that a reclaimed slur is often ascribed to oneself, we propose to use a supporting hypothesis that indicates if text is self-directed.\footnote{Of course there are counterexamples to this rule, where reclaimed slurs are directed to others and not oneself. However, as long this approximation, as crude as it may be, helps to reduce false positives, it is a useful approximation.}
If the model predicts self-directedness a potential prediction of \textit{hate speech} is overridden to \textit{not-hate speech}. 

\subsection{Catching Dehumanizing Comparisons (CDC)}
\label{subsec:catching-dehumanizing-comparisons}

One way of expressing hate towards a group and dehumanizing said group is to draw unflattering comparisons with animals. Such comparisons tend to be missed by hate speech detection systems, since the use of hateful or aggressive words is not needed to convey the hateful message. In HateCheck, this phenomenon is subsumed under \enquote{Implicit derogation}.
Standard zero-shot prediction obtains a moderately good accuracy of 89.3\%. 
We test if false negatives can be caught with a three-step combination of supporting hypotheses:
(1) use the supporting hypotheses of FBT to predict if a protected group is mentioned in text $t$, 
(2) predict if $t$ has a negative sentiment, and
(3) predict if $t$ has is about animals typically used when making dehumanizing comparisons (such as insects, rats, or monkeys). 
If all conditions are met, override a prediction of \textit{not-hate speech} to \textit{hate speech}.

\section{Experiments}
\label{sec:experiments}

We use the same model and adopt the entailment threshold of 0.5 from Section \ref{sec:evaluating-standard-0shot} for the main and all supporting hypotheses.
Further, we take the hypothesis leading to the highest accuracy in Section \ref{sec:evaluating-standard-0shot} as the main hypothesis.

Since the main hypothesis in our experiments is chosen for maximum accuracy on HateCheck (based on the experiment in Section \ref{sec:evaluating-standard-0shot}) and the strategies developed are based on an error analysis on HateCheck, the overall system might be overfitting on this specific dataset. An evaluation on this dataset might thus lead to results that overestimate a potential positive effect of the proposed strategies. We therefore also evaluate on ETHOS as an ``unseen'' dataset.

ETHOS does not refer to protected groups in its definition and annotation of hate speech, but instead to protected characteristics. Thus, in the hypotheses for FBT we replace protected groups with the protected characteristics listed in Table \ref{tab:list-of-hypotheses}.

\subsection{Results}

\paragraph{HateCheck} The bottom row \textit{Overall} in Table \ref{tab:full-error-analysis} shows the results for the proposed strategies and their combination on the HateCheck dataset. Each strategy leads to an improvement in accuracy. But while FBT and FCS lead to large increases, FRS and CDC only lead to minor increases. Combining all proposed strategies leads to an increase in accuracy of 7.9pp.

\begin{table}[t]
\centering
\small
\resizebox{\textwidth / 2}{!}{%
\begin{tabular}{lrr}
\hline
strategies & accuracy (\%) & $\Delta$ \\
\hdashline
(ETHOS) SVM & 66.4 & - \\
(ETHOS) BERT & 80.0 & - \\
(ETHOS) DistilBERT & 80.4 & - \\
\hdashline
``That contains hate speech.'' & 69.6 & +0.0 \\
\hdashline
FBT (TG) & 75.5 & +5.9 \\
FBT (TC) & 78.7 & +9.1 \\
FCS & 69.6 & +0.0 \\
FRS & 71.3 & +1.7 \\
CDC (TC) & 69.5 & -0.1 \\
\hdashline
FBT (TC) + FCS & 78.7 & +9.1 \\
FBT (TC) + FRS & 79.7 & +10.1 \\
FCS + FRS & 71.3 & +1.7 \\
FBT (TC) + FCS + FRS & 79.7 & +10.1 \\
CDC (TC) + FBT (TC) + FCS + FRS & 79.6 & +10.0 \\
\hline
\end{tabular} 
}
\caption{Accuracy scores on ETHOS. The three top rows show baselines computed by \citet{mollas_ethos_2022}. TG refers to using \textit{target groups} to implement FBT and TC refers to using \textit{target characteristics} for FBT.}
\label{tab:full-results-ethos}
\end{table}

\paragraph{ETHOS} The results of evaluating the same strategies on ETHOS \cite{mollas_ethos_2022} are given in Table \ref{tab:full-results-ethos}. As additional baselines compared to zero-shot prediction using just one hypothesis, we include the performance of three models trained on ETHOS by \citet{mollas_ethos_2022}.

The combination of all strategies leads to a increase of 10.0pp, which is an even greater increase than on HateCheck.
However, the gains are more unevenly distributed across the proposed strategies. 
Filtering by target characteristics alone leads to an increase of 9.1pp. Filtering reclaimed slurs still has a positive effects of 1.7pp. However, filtering counterspeech, the best performing strategy on HateCheck, does not have any effect at all. And catching dehumanizing comparisons even reduces performance by 0.1pp.

The comparison to the baselines provided by \citet{mollas_ethos_2022} shows that zero-shot prediction using the hypothesis ``That contains hate speech.'' already outperforms a trained SVM by more than 3pp but still underperforms the fine-tuned BERT by more than 10pp. However, applying the proposed strategies almost closes the gap to the fine-tuned models.

\subsection{Analysis of Affected Functionalities}

We analyse if the observed performance gains actually stem from improvements on the functionalities targeted by the proposed strategies. Table \ref{tab:full-error-analysis} shows for each functionality how it was affected by each strategy. 

\paragraph{Filtering by Target} The results for filtering by target show dramatic accuracy increases for HateCheck functionalities containing abuse and profanity that is not targeted at a protected group. These are exactly the functionalities this strategy aimed at. 
The performance for spelling variations and implicit derogation decreases slightly. This can be explained by the model failing to correctly recognize spelling variations of target groups and by the fact that the target group might only be implied in implicit derogation leading to false negatives.

\paragraph{Filtering Counterspeech}
The counterspeech filter increases the accuracy of the respective functionality from 0\% to 100\%. Thus, detecting quoted hate speech as well as detecting the stance towards the quote worked exactly as intented on HateCheck.

\paragraph{Filtering Reclaimed Slurs} The functionality with the largest gains when filtering reclaimed slurs is ``reclaimed slurs'', showing that the strategy works as intended. However, the performance increase of this method is not as high as for example filtering by target. The functionality ``non-hateful use of profanity'' also benefits from this strategy. We assume that such uses of profanity often are also not directed at other people and thus sometimes predicted to be directed at oneself. This is a beneficial side-effect of the strategy.

\paragraph{Catching Dehumanizing Comparisons} This strategy only leads to minor a minor overall improvement of 0.2pp. We observe no effect on the targeted functionality, but a small positive effect on F7, ``Hate expressed using slur'', which could indicate that the model associates slurs with negatively coded animals. Additionally, the strategy has minor positive effects on functionalities that contain spelling variations.

\section{Discussion}

\paragraph{Supporting Hypotheses} The performance of NLI-strategies largely depends on the accuracy of the supporting hypotheses. Testing the accuracy of each supporting hypothesis is not always possible, since annotated data for the predicted aspect of the input text might not be available.
Indeed, one of the strengths of our approach is that it can use aspects for which no annotated data exists.
Another uncertainty lies in the formulation of supporting hypotheses. A suboptimal formulation of supporting hypotheses negatively affects the overall results. 
By using annotated targets in HateCheck and inferring stance labels as well as self-directedness from HateCheck functionalities, we compute and compare the accuracy of multiple supporting hypothesis formulations.
The results (in Appendix \ref{appsec:eval-aux-hypos}) show that testing for the presence of a target group mostly leads to accuracies above 90\%, independent of the specific formulation. Detecting the outer stance towards the inner text obtained a perfect accuracy of 100\% and testing for self-directedness leads to low accuracies, which are probably partly due to faulty label inferences from functionalities.
Overall, the results indicate that the supporting hypotheses provide reliable information.

\begin{figure}[t]
\center
\includegraphics[width=1\linewidth]{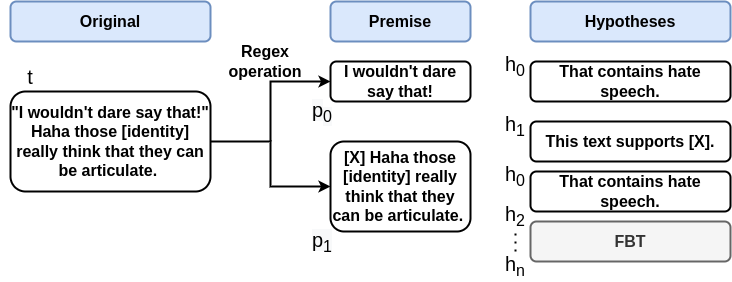}
\caption{Counterspeech filter adjusted for detecting hate speech where quotes are present but the hate speech is outside of the quote - i.e. in the outer text.}
\label{fig:counterspeech-filter-fbt}
\end{figure}

\begin{table}[ht]
\centering
\small
\begin{tabular}{lrr}
\hline
strategy & F20 & overall \\ 
\hdashline
No strategy & 0.0 & 79.4 \\
FCS & +100 & +4.6 \\
\hdashline
FCS$_{p_1}$ & +0.0 & +0.0 \\
FCS$_{p_1 FBT}$ & +6.9 & +0.3 \\
\hline
\end{tabular}
\caption{Evaluation of FCS variants. The two bottom rows display the variants adjusted for detecting Hate Speech in $p_1$. The functionality F20 contains \textit{Denouncements of hate that quote it}. The scores are given in accuracy (\%) and change in accuracy compared to \textit{No strategy}.}
\label{tab:FCS-follow-up}
\end{table}

\paragraph{Generality} There are two ways in which the proposed strategies might not generalize. 
First, the strategies might be specific to the model used for the experiments. In order to answer this question, repeating the experiments with other NLI models will be necessary. 
Second, the strategies might be specific to HateCheck and not generalize to other datasets, since they specifically target HateCheck functionalities. The evaluation on ETHOS shows that this is generally not case, since the results for the best strategy combination on ETHOS even exceed the results for the best combination on HateCheck. 

While our experiments did not show problems in generalization, we can imagine the following weakness for the FCS strategy: 
Given an input text $t$ that contains a quote and hate speech, where the hate speech does not occur inside of the quotes, the current FCS strategy would fail, since it only detects hate speech in $p_0$, that is inside the quotes. Such an example is given in Figure \ref{fig:counterspeech-filter-fbt}.

The obvious solution to avoid this problem is to not only apply the main hypothesis $h_0$ on the premise $p_0$ but also on premise $p_1$, which contains the text outside the quotes. 
In a follow-up experiment we implement this modified strategy (FCS$_{p_1}$) and evaluated it on HateCheck.
Note, that since such a case is not covered by HateCheck or ETHOS there is no increase in accuracy to be expected - we can only test if accounting for this case leads to a decrease in accuracy through unwanted side effects. 

The results, displayed in Table \ref{tab:FCS-follow-up} show that this modification removes all the gains obtained through FBT. 
We assume that this is due to the fact that the counterspeech often also conveys strong negative emotions that are mistaken by the model for hate speech.

We further test if this problem can be alleviated by applying the FBT strategy if hate speech is detected in $p_1$ (i.e. outside of the quotes) as depicted in Figure \ref{fig:counterspeech-filter-fbt}. 
The results in Table \ref{tab:FCS-follow-up} (row FCS$_{p_1FBT}$) show that additionally applying FBT only recovers a fraction of the positive effect of FCS. 
We assume that this is due to counterspeech including or being associated with target groups. 
Thus, further research that investigates how the problem can be alleviated is needed.

\paragraph{Efficiency} In the proposed setup each new hypothesis necessitates an additional forward pass, which means that the computational cost linearly increases with adding new hypotheses. This leads to a difficult trade-off between accuracy and efficiency. 
A possible solution was recently proposed by \citet{muller_few-shot_2022}, who embed premises and hypotheses independently, thereby keeping the computational cost during inference time with respect to the number of hypotheses constant. 

\paragraph{Prerequisites of FBT} FBT presumes that the target groups or characteristics are known beforehand. This prerequisite is unproblematic when using FBT to detect hate speech against well known targets of hate speech or discrimination. However, it makes this method unsuitable for tasks such as vulnerable group identification \cite{mossie_vulnerable_2020}.

\paragraph{Flexibility} 
Single hypotheses or entire strategies can be easily added to or removed from a system. This modularity makes the approach easily adjustable to different scenarios or use cases. 
For example, if precision is the main concern, the \textit{catching dehumanizing comparisons} can be dropped and if recall is the main concern, filters can be removed. 
Instead of adding or removing strategies, it is also possible to manipulate the precision-recall trade-off by adjusting confidence thresholds for particular hypotheses.

\section{Conclusion}

In this work, we combine hypotheses to create more accurate NLI-based zero-shot hate speech detection systems. 
Specifically, we develop four simple strategies, \textit{filtering by target}, \textit{filtering counter speech}, \textit{filtering reclaimed-slurs}, and \textit{catching dehumanizing comparisons}, that target specific model weaknesses.
We evaluate the strategies on HateCheck, which served as the basis for developing these strategies, and on ETHOS, which acts as an \enquote{unseen} dataset.
The NLI-based zero-shot baseline already outperforms fine-tuned models on HateCheck and beats an SVM baseline on ETHOS.
Using all four proposed strategies leads to a further performance increase of 7.9\% on HateCheck and 10.0\% on ETHOS.
However, the contribution of the strategies to the performance increases varies, with \textit{catching dehumanizing comparisons} even having a small negative effect on the accuracy on ETHOS.

The proposed approach is simple and modular making it easy to implement and adjust to different scenarios.

In future work, we plan to evaluate such strategies in a multi-lingual setup and
in a few-shot scenario.
Further, this works leads to the question how effective supporting hypotheses could be searched and generated automatically.

\section*{Acknowledgements}

We thank Chantal Amrhein, Jonathan Schaber, Aneta Zuber, and the anonymous reviewers for their helpful feedback. This work was funded by the University of Zurich Research Priority Program (project \enquote{URPP Digital Religion(s)}\footnote{\url{https://www.digitalreligions.uzh.ch/en.html}}). 

\section*{Ethical Considerations}

The goal of this article is to contribute to the development of sophisticated hate speech detection methods, and thus to contribute to an online environment that is less hateful. However, we can imagine multiple ways how such methods, and our proposed approach in particular, can lead to harm: 
(1) Deploying the exact system that we propose would lead to not detecting hate speech against protected groups that are not explicitly included in the two datasets we worked with and thus not covered by the FBT-method. Thus, before deploying such a method, careful consideration of which protected groups or group characteristics are covered is needed. 
(2) Overconfident claims about the accuracy of hate speech detection methods could lead to the false impression that content moderation can be left to automatic methods with no human intervention. 
(3) Hate speech detection in general is prone to misuse and repurposing in order to prohibit other kinds of speech. 
Detecting if a text revolves around a protected group could be misused to detect and prohibit important discussion around topics connected to a protected group. 

However, we believe that a decomposition of hate speech into more specific aspects is important for more accurate, interpretable and modular hate speech detection methods \cite{khurana_hate_2022}. Thus, detecting such components of hate speech is also strongly beneficial for effective content moderation and a less hateful online environment.

\bibliography{acl}
\bibliographystyle{acl_natbib}
\newpage 

\appendix

\section{Zero-Shot Results: Comparing Hypotheses}
\label{appsec:zero-shot-results-compare-hypos}

Table \ref{tab:compare-hypotheses-full}, the extended version of Table \ref{tab:compare-hypotheses}, contains all results for comparing hypotheses for zero-shot hate speech detection on HateCheck. \enquote{average:$<$ \texttt{expression}$>$} refers to the average accuracy of all hypotheses containing \texttt{expression}. The highest accuracy is in bold. 

\begin{table}[h!]
\centering
\small
\begin{tabular}{lr}
\hline
hypothesis & accuracy (\%) \\ 
\hdashline
Containing hate speech. & 74.7 \\
Contains hate speech. & 78.6 \\
Hate speech. & 72.9 \\
Hateful. & 71.8 \\
\hdashline
It contains hate speech. & 78.7 \\
It is hateful. & 75.0 \\
It contains hate speech. & 78.7 \\
It is hate speech. & 70.8 \\
It is hateful. & 75.0 \\
That contains hate speech. & \textbf{79.4} \\
That contains hateful content. & 78.0 \\
That example contains hateful content. & 77.8 \\
That example is hate speech. & 66.6 \\
That example is hateful. & 76.8 \\
That is hateful. & 66.6 \\
That text contains hate speech. & 78.8 \\
That text contains hateful content. & 78.6 \\
That text is hate speech. & 69.2 \\
That text is hateful. & 77.2 \\
This contains hate speech. & 79.1 \\
This contains hateful content. & 78.2 \\
This example contains hate speech. & 77.3 \\
This example contains hateful content. & 77.8 \\
This example is hate speech. & 67.2 \\
This example is hateful. & 77.4 \\
This is hateful. & 70.6 \\
This text contains hate speech. & 78.8 \\
This text contains hateful content. & 78.3 \\
This text is hate speech. & 69.5 \\
This text is hateful. & 78.7 \\
\hdashline
average: It & 74.8 \\
average: This & 75.7 \\
average: That & 74.9 \\
average: hateful & 75.8 \\
average: hateful content & 78.1 \\
average: hate speech & 74.5 \\
average: example & 74.4 \\
average: text & 76.1 \\
average: is & 73.9 \\
average: contain & 78.2 \\
\hdashline
SiftNinja & 33.2 \\
BERT fine-tuned on \citet{davidson_automated_2017} & 60.2 \\
BERT fine-tuned on \citet{founta_large_2018} & 63.2 \\
Google-Jigsaw & 76.6 \\
\hline
\end{tabular}
\caption{Full evaluation of hypotheses, that claim hate speech exists in the input text, on HateCheck.}
\label{tab:compare-hypotheses-full}
\end{table}

\section{Evaluating Supporting Hypotheses}
\label{appsec:eval-aux-hypos}

\subsection{Target Groups and Target Characteristics}

Each example in HateCheck which mentions a protected group or revolves around a protected group is annotated with said group. If no group is targeted the example is annotated with an empty string. By using these annotations as labels, we can create a binary classification task for each protected group: for detecting a mention of a specific protected group $x$, we convert label $x$ to $1$ and all other labels (i.e. all other protected groups) to $0$. 

We use the same model and as in the previous zero-shot experiments for evaluation and test the performance for detecting mentions for all protected groups in HateCheck. We additionally test the detection of the supercategory \textit{queer people} covering the two protected groups \textit{gay people} and \textit{transgender people} in HateCheck. When testing if a text revolves around gender, we treat both \textit{women} and \textit{transgender people} as positive classes and all other protected groups as negative classes. While this mapping obviously can result in incorrect labels (a text can be about gender even if another group is targeted), we assume that it holds true for examples in the HateCheck dataset.

Table \ref{tab:results-black-people} shows the results for detecting if \textit{black people} are mentioned, Table \ref{tab:results-muslim-people} for mentions of \textit{Muslims}, Table \ref{tab:results-immigrants} for mentions of immigrants, Table \ref{tab:results-disabled-people} for mentions of \textit{disabled people}, Table \ref{tab:results-gay-people} for mentions of \textit{gay people}, Table \ref{tab:results-transgender-people} for mentions of \textit{transgender people}, Table \ref{tab:results-queer-people} for mentions of \textit{queer people}, and Table \ref{tab:results-gender} for detecting if a text is about \textit{gender}.

The results show that in many cases the detection of a mentioned group is surprisingly accurate. The difference in accuracy between the best performing hypothesis and the worst performing hypothesis does not exceed 12\%. This is a similar range to the differences found between hypotheses when testing if a text contains hate speech (see Table \ref{tab:compare-hypotheses} and Table \ref{tab:compare-hypotheses-full}). However, when looking at $F_1$ scores the differences are much larger, with more general terms, such as \textit{faith} or \textit{ethnicity} preforming worse than the specific terms \textit{Muslims} and \textit{black people}.

Detecting if a text revolves around gender performs worst, compared to detecting other protected groups or characteristics. This is mostly due to low precision scores. We assume that this is a consequence of sexual orientation (\textit{gay people}) being closely associated in the embedding space with \textit{gender} and thus leading to false positives. 

\begin{table*}[t]
\small
\centering
\begin{tabular}{l|rrrr}
hypothesis & \multicolumn{1}{l}{accuracy (\%)} & \multicolumn{1}{l}{$\downarrow$ $F_1$ (\%)} & \multicolumn{1}{l}{recall (\%)} & \multicolumn{1}{l}{precision (\%)} \\ \hline
That example is about black people.     & 97.9 & 92.0 & 93.8 & 90.2 \\
This example is about black people.     & 97.6 & 90.9 & 94.0 & 88.0 \\
That text is about black people.        & 96.4 & 87.1 & 93.2 & 81.8 \\
That is about black people.             & 95.8 & 85.0 & 92.5 & 78.7 \\
This text is about black people.        & 95.2 & 83.6 & 94.4 & 75.0 \\
This is about black people.             & 95.0 & 83.0 & 93.6 & 74.5 \\
That example is about people of colour. & 94.3 & 80.7 & 92.7 & 71.4 \\
That example is about race.             & 94.0 & 79.8 & 91.3 & 70.9 \\
This example is about people of colour. & 93.8 & 79.3 & 92.3 & 69.5 \\
That is about race.                     & 94.6 & 77.5 & 72.6 & 83.1 \\
This example is about race.             & 90.9 & 72.3 & 91.7 & 59.6 \\
That text is about people of colour.    & 89.8 & 70.6 & 94.2 & 56.4 \\
That example is about ethnicity.        & 90.5 & 69.8 & 85.3 & 59.1 \\
This is about race.                     & 88.9 & 67.6 & 89.4 & 54.4 \\
That text is about race.                & 87.8 & 66.3 & 92.3 & 51.7 \\
That is about people of colour.         & 87.4 & 65.9 & 93.8 & 50.8 \\
This text is about race.                & 86.5 & 64.5 & 94.6 & 48.9 \\
This text is about people of colour.    & 84.9 & 62.1 & 96.1 & 45.9 \\
This example is about ethnicity.        & 85.6 & 62.1 & 91.1 & 47.1 \\
This is about ethnicity.                & 85.4 & 58.1 & 78.4 & 46.2 \\
That text is about ethnicity.           & 81.5 & 56.1 & 91.3 & 40.5 \\
This text is about ethnicity.           & 80.2 & 55.0 & 93.6 & 38.9 \\
This is about people of colour.         & 74.8 & 49.4 & 95.2 & 33.3 \\
That is about ethnicity.                & 87.0 & 28.4 & 19.9 & 49.7
\end{tabular}
\caption{Results for supporting hypotheses aimed at detecting mentions of black people. The hypotheses are sorted by macro $F_1$-score in descending order. Note, that some of the hypotheses listed use broader terms (\enquote{people of colour}, \enquote{race}, \enquote{ethnicity}) that should also detect the mentions of other target groups. However, in the context of HateCheck, we can only test the detection of mentioning black people.}
\label{tab:results-black-people}
\end{table*}

\begin{table*}[t]
\small
\centering
\begin{tabular}{l|rrrr}
hypothesis & \multicolumn{1}{l}{accuracy (\%)} & \multicolumn{1}{l}{$\downarrow$ $F_1$ (\%)} & \multicolumn{1}{l}{recall (\%)} & \multicolumn{1}{l}{precision (\%)} \\ \hline
That example is about Muslims.       & 98.3 & 93.1 & 90.7 & 95.6 \\
This example is about Muslims.       & 98.1 & 92.7 & 90.7 & 94.8 \\
This text is about Muslims.          & 98.1 & 92.6 & 90.3 & 95.0 \\
That text is about Muslims.          & 98.0 & 92.3 & 90.3 & 94.4 \\
That example is about Muslim people. & 98.0 & 92.0 & 90.7 & 93.4 \\
This example is about Muslim people. & 97.9 & 92.0 & 90.9 & 93.0 \\
This is about Muslims.               & 97.8 & 91.6 & 90.7 & 92.4 \\
This text is about Muslim people.    & 97.8 & 91.5 & 90.7 & 92.2 \\
That text is about Muslim people.    & 97.7 & 91.1 & 90.1 & 92.2 \\
This example is about religion.      & 97.7 & 90.3 & 83.3 & 98.5 \\
This text is about religion.         & 97.5 & 89.7 & 83.9 & 96.4 \\
This is about Muslim people.         & 97.3 & 89.6 & 89.9 & 89.3 \\
That text is about religion.         & 97.5 & 89.5 & 82.9 & 97.3 \\
That is about Muslims.               & 97.2 & 89.1 & 88.6 & 89.6 \\
That example is about religion.      & 97.3 & 88.5 & 80.2 & 98.7 \\
That is about Muslim people.         & 96.8 & 87.8 & 89.7 & 85.9 \\
This is about religion.              & 96.4 & 84.5 & 74.8 & 97.1 \\
This example is about faith.         & 95.0 & 78.2 & 69.2 & 89.8 \\
This text is about faith.            & 95.0 & 78.1 & 69.0 & 90.0 \\
That text is about faith.            & 94.4 & 74.2 & 62.4 & 91.5 \\
That example is about faith.         & 93.4 & 68.9 & 56.6 & 88.1 \\
This is about faith.                 & 92.0 & 63.3 & 52.9 & 78.8 \\
That is about religion.              & 89.6 & 34.5 & 21.1 & 95.3 \\
That is about faith.                 & 88.3 & 21.9 & 12.6 & 82.4
\end{tabular}
\caption{Results for supporting hypotheses aimed at detecting mentions of Muslim people. Note, that some of the hypotheses listed use broader terms (\enquote{faith}, \enquote{religion}) that should detect other target groups too. However, in the context of HateCheck their applicability is restricted to Muslim people, since no other religion occurs in HateCheck.}
\label{tab:results-muslim-people}
\end{table*}

\begin{table*}[t]
\small
\centering
\begin{tabular}{l|rrrr}
hypothesis & \multicolumn{1}{l}{accuracy (\%)} & \multicolumn{1}{l}{$\downarrow$ $F_1$ (\%)} & \multicolumn{1}{l}{recall (\%)} & \multicolumn{1}{l}{precision (\%)} \\ \hline
That example is about immigrants.      & 97.8 & 91.2 & 92.4 & 89.9 \\
This example is about immigrants.      & 97.7 & 90.8 & 92.4 & 89.2 \\
That is about immigrants.              & 97.2 & 88.8 & 89.2 & 88.4 \\
That text is about immigrants.         & 97.0 & 88.6 & 92.2 & 85.2 \\
This text is about immigrants.         & 96.3 & 86.4 & 93.7 & 80.1 \\
This is about immigrants.              & 96.4 & 86.2 & 91.6 & 81.4 \\
This text is about national origin.    & 77.7 & 42.2 & 65.4 & 31.1 \\
That text is about national origin.    & 78.0 & 41.3 & 62.4 & 30.9 \\
This is about national origin.         & 83.4 & 37.0 & 39.3 & 35.0 \\
That example is about national origin. & 81.4 & 30.0 & 32.2 & 28.1 \\
This example is about national origin. & 79.8 & 30.0 & 34.8 & 26.3 \\
That is about national origin.         & 86.5 & 24.6 & 17.7 & 40.2
\end{tabular}
\caption{Results for supporting hypotheses aimed at detecting mentions of immigrants. }
\label{tab:results-immigrants}
\end{table*}

\begin{table*}[t]
\small
\centering
\begin{tabular}{l|rrrr}
hypothesis & \multicolumn{1}{l}{accuracy (\%)} & \multicolumn{1}{l}{$\downarrow$ $F_1$ (\%)} & \multicolumn{1}{l}{recall (\%)} & \multicolumn{1}{l}{precision (\%)} \\ \hline
That example is about disabled people. & 98.4 & 93.7 & 90.3 & 97.3 \\
This example is about disabled people. & 98.4 & 93.5 & 90.1 & 97.1 \\
This text is about disabled people.    & 98.0 & 92.3 & 91.3 & 93.2 \\
That example is about disability.      & 97.9 & 91.9 & 92.4 & 91.4 \\
That text is about disabled people.    & 97.9 & 91.3 & 87.0 & 96.1 \\
This example is about disability.      & 97.7 & 91.3 & 93.0 & 89.6 \\
That is about disabled people.         & 97.7 & 90.9 & 87.2 & 94.8 \\
This is about disabled people.         & 97.4 & 89.9 & 90.3 & 89.5 \\
That text is about disability.         & 96.6 & 87.6 & 91.5 & 84.1 \\
That is about disability.              & 96.6 & 86.0 & 80.4 & 92.4 \\
This is about disability.              & 94.8 & 82.0 & 91.7 & 74.1 \\
This text is about disability.         & 94.5 & 81.6 & 94.4 & 71.9
\end{tabular}
\caption{Results for supporting hypotheses aimed at detecting mentions of disabled people.}
\label{tab:results-disabled-people}
\end{table*}

\begin{table*}[t]
\small
\centering
\begin{tabular}{l|rrrr}
hypothesis & \multicolumn{1}{l}{accuracy (\%)} & \multicolumn{1}{l}{$\downarrow$ $F_1$ (\%)} & \multicolumn{1}{l}{recall (\%)} & \multicolumn{1}{l}{precision (\%)} \\ \hline
That example is about gay people. & 99.1 & 96.7 & 94.2 & 99.4 \\
This example is about gay people. & 99.0 & 96.6 & 94.0 & 99.2 \\
This text is about gay people.    & 98.8 & 95.9 & 92.9 & 99.0 \\
This is about gay people.         & 98.8 & 95.8 & 92.6 & 99.2 \\
That text is about gay people.    & 98.5 & 94.6 & 90.0 & 99.6 \\
That is about gay people.         & 98.4 & 94.4 & 90.4 & 98.8
\end{tabular}
\caption{Results for supporting hypotheses aimed at detecting mentions of gay people. }
\label{tab:results-gay-people}
\end{table*}

\begin{table*}[t]
\small
\centering
\begin{tabular}{l|rrrr}
hypothesis & \multicolumn{1}{l}{accuracy (\%)} & \multicolumn{1}{l}{$\downarrow$ $F_1$ (\%)} & \multicolumn{1}{l}{recall (\%)} & \multicolumn{1}{l}{precision (\%)} \\ \hline
That example is about transgender people. & 99.0 & 95.8 & 92.7 & 99.1 \\
That text is about transgender people.    & 99.0 & 95.6 & 92.2 & 99.3 \\
This text is about transgender people.    & 98.9 & 95.6 & 92.9 & 98.4 \\
This example is about transgender people. & 98.9 & 95.4 & 92.2 & 98.8 \\
That is about transgender people.         & 98.8 & 94.9 & 92.0 & 97.9 \\
This is about transgender people.         & 98.7 & 94.6 & 92.2 & 97.0
\end{tabular}
\caption{Results for supporting hypotheses aimed at detecting mentions of transgender people.}
\label{tab:results-transgender-people}
\end{table*}

\begin{table*}[t]
\small
\centering
\begin{tabular}{l|rrrr}
hypothesis & \multicolumn{1}{l}{accuracy (\%)} & \multicolumn{1}{l}{$\downarrow$ $F_1$ (\%)} & \multicolumn{1}{l}{recall (\%)} & \multicolumn{1}{l}{precision (\%)} \\ \hline
This is about queer people.         & 94.3 & 88.6 & 81.3 & 97.5 \\
That example is about queer people. & 93.7 & 87.1 & 77.9 & 98.8 \\
This example is about queer people. & 93.2 & 85.9 & 76.2 & 98.5 \\
This text is about queer people.    & 93.1 & 85.7 & 76.4 & 97.5 \\
That is about queer people.         & 92.4 & 84.0 & 73.5 & 98.2 \\
That text is about queer people.    & 91.7 & 82.3 & 70.8 & 98.4
\end{tabular}
\caption{Results for supporting hypotheses aimed at detecting mentions of queer people, which in HateCheck corresponds to the categories \textit{gay people} and \textit{transgender people}.}
\label{tab:results-queer-people}
\end{table*}

\begin{table*}[t]
\small
\centering
\begin{tabular}{l|rrrr}
hypothesis & \multicolumn{1}{l}{accuracy (\%)} & \multicolumn{1}{l}{$\downarrow$ $F_1$ (\%)} & \multicolumn{1}{l}{recall (\%)} & \multicolumn{1}{l}{precision (\%)} \\ \hline
This example is about women. & 97.2 & 90.2 & 94.1 & 86.6 \\
That example is about women. & 97.2 & 90.1 & 94.1 & 86.5 \\
That is about women.         & 96.2 & 86.6 & 88.6 & 84.6 \\
This text is about women.    & 95.9 & 86.1 & 93.9 & 79.5 \\
That text is about women.    & 95.7 & 85.3 & 91.6 & 79.8 \\
This is about women.         & 94.8 & 83.0 & 92.7 & 75.0
\end{tabular}
\caption{Results for supporting hypotheses aimed at detecting mentions of women.}
\label{tab:results-women}
\end{table*}

\begin{table*}[t]
\small
\centering
\begin{tabular}{l|rrrr}
hypothesis & \multicolumn{1}{l}{accuracy (\%)} & \multicolumn{1}{l}{$\downarrow$ $F_1$ (\%)} & \multicolumn{1}{l}{recall (\%)} & \multicolumn{1}{l}{precision (\%)} \\ \hline
That example is about gender. & 90.0 & 81.8 & 85.8 & 78.2 \\
This example is about gender. & 89.0 & 80.5 & 87.0 & 74.9 \\
That text is about gender.    & 88.4 & 79.6 & 86.8 & 73.5 \\
This text is about gender.    & 87.9 & 79.3 & 88.9 & 71.5 \\
This is about gender.         & 87.1 & 75.1 & 74.7 & 75.5 \\
That is about gender.         & 81.6 & 52.9 & 39.5 & 79.8
\end{tabular}
\caption{Results for supporting hypotheses aimed at detecting texts concerning gender, which in HateCheck corresponds to the categories \textit{transgender people}, and \textit{women}.}
\label{tab:results-gender}
\end{table*}


\subsection{Self-Directedness}
\label{appsubsec:self-directedness}

Evaluating the accuracy of detecting self-directedness is difficult, because there exist no labels in HateCheck that could be used as a ground truth. 

One possibility, that follows the motivation for introducing the FRS-method, is to treat all examples of functionality F9 (\enquote{reclaimed\_slur}) as self-directed and examples of all other functionalities as not self-directed. We conducted this experiment. The results are given in Table \ref{tab:results-self-directed}. 
However, one should keep in mind that this disregards that reclaimed slurs can be used in a not-self directed manner and that other functionalities, such as functionality F11 \textit{non-hateful use of profanity}, might contain examples of self-directed speech.

\begin{table*}[t]
\small
\centering
\begin{tabular}{l|rrrr}
hypothesis & \multicolumn{1}{l}{accuracy (\%)} & \multicolumn{1}{l}{$\downarrow$ $F_1$ (\%)} & \multicolumn{1}{l}{recall (\%)} & \multicolumn{1}{l}{precision (\%)} \\ \hline
That text is about myself.    & 97.4 & 38.5 & 37.0 & 40.0 \\
This text is about myself.    & 96.4 & 33.7 & 42.0 & 28.1 \\
That is about myself.         & 97.3 & 33.3 & 30.9 & 36.2 \\
This is about myself.         & 97.0 & 31.3 & 30.9 & 31.6 \\
That example is about myself. & 96.2 & 31.2 & 39.5 & 25.8 \\
This example is about myself. & 95.9 & 31.1 & 42.0 & 24.6 \\
This text is about us.        & 85.1 & 16.8 & 69.1 & 9.6  \\
That text is about us.        & 89.4 & 16.2 & 46.9 & 9.8  \\
That is about us.             & 88.3 & 14.8 & 46.9 & 8.8  \\
This example is about us.     & 77.4 & 12.8 & 76.5 & 7.0  \\
That example is about us.     & 75.8 & 12.1 & 76.5 & 6.6  \\
This is about us.             & 73.9 & 10.3 & 69.1 & 5.6 
\end{tabular}
\caption{Results for supporting hypotheses aimed detecting if a text is self-directed.}
\label{tab:results-self-directed}
\end{table*}

\subsection{Counterspeech}
\label{appsubsec:counterspeech}

We perform a simple evaluation of the supporting hypothesis that predicts the stance of an outer text towards its quoted inner text (see Section \ref{subsec:filtering-counterspeech} for an explanation) using only functionality F20 (\textit{Denouncements of hate that quote it}) as an evaluation set. We treat stance detection here as a binary task with the labels \textit{is\_for} or \textit{is\_against}. 

How these labels are mapped onto NLI labels depends on the specific hypothesis. If the hypothesis claims that the outer text supports the quoted text, then \textit{is\_for} is mapped to \textit{entailment} and \textit{is\_against} is mapped to \textit{contradiction}. Conversely, if the hypothesis claims that the outer text denounces the quoted text, then \textit{is\_for} is mapped to \textit{contradiction} and \textit{is\_against} is mapped to \textit{entailment}.

We test various formulations including the verbs \enquote{supports $[$X$]$} and \enquote{is for $[$X$]$}. 
The results are given in Table \ref{tab:results-stance}. Since all examples in this category are considered hate speech, that is denounced by the outer text, the true label is always \textit{is\_against}. We only report accuracy, since there can be no false positives and true negatives, which makes precision and recall lose its usefulness.

\begin{table*}[h]
\small
\centering
\begin{tabular}{l|r}
hypothesis & \multicolumn{1}{l}{accuracy (\%)} \\ 
\hline
This text supports [X].    & 100.0 \\
This supports [X].         & 100.0 \\
That supports [X].         & 100.0 \\
This example supports [X]. & 91.9  \\
That example supports [X]. & 91.9  \\
That text supports [X].    & 85.5  \\
This text is for [X].      & 69.4  \\
This is for [X].           & 50.9  \\
That is for [X].           & 46.8  \\
That text is for [X].      & 38.7  \\
This example is for [X].   & 18.5  \\
That example is for [X].   & 0.0   \\
\end{tabular}
\caption{Results for supporting hypotheses aimed at detecting the stance of an outer text $p_1$ towards its inner, quoted text $p_0$. Precision, recall and $F_1$-score are omitted, since with only positive test examples no false positives and true negatives are possible.}
\label{tab:results-stance}
\end{table*}

\end{document}